\documentclass[letterpaper, 10 pt, conference]{ieeeconf}  

\IEEEoverridecommandlockouts                              

\usepackage{graphics} 
\usepackage{amsmath} 
\usepackage{amssymb}  
\usepackage{xcolor}
\usepackage{graphicx}
\usepackage{hyperref}
\usepackage[capitalize]{cleveref}
\usepackage{xcolor}

\usepackage{tabularx}
\usepackage{makecell}
\usepackage{booktabs} 

\DeclareMathOperator{\diag}{diag}

\title{\LARGE \bf
Adaptive Nonlinear Model Predictive Control for a Real-World Labyrinth Game
}

\author{Johannes Gaber$^{1}$, Thomas Bi$^{1}$, Raffaello D'Andrea$^{1}$
	\thanks{$^{1}$The authors are with the Institute for Dynamic Systems and Control, ETH Zurich, Switzerland, {\tt\small jgaber@student.ethz.ch}}%
}
\Crefname{figure}{Fig.}{Fig.}

\begin{document}

\maketitle
\thispagestyle{empty}
\pagestyle{empty}


\begin{abstract}

We present a nonlinear non-convex model predictive control approach to solving a real-world labyrinth game. We introduce adaptive nonlinear constraints, representing the non-convex obstacles within the labyrinth. Our method splits the computation-heavy optimization problem into two layers; first, a high-level model predictive controller which incorporates the full problem formulation and finds pseudo-global optimal trajectories at a low frequency. Secondly, a low-level model predictive controller that receives a reduced, computationally optimized version of the optimization problem to follow the given high-level path in real-time. Further, a map of the labyrinth surface irregularities is learned. Our controller is able to handle the major disturbances and model inaccuracies encountered on the labyrinth and outperforms other classical control methods.

\end{abstract}
\section{Introduction}
\label{introduction}

\begin{figure}
    \centering
    \includegraphics[height=4cm]{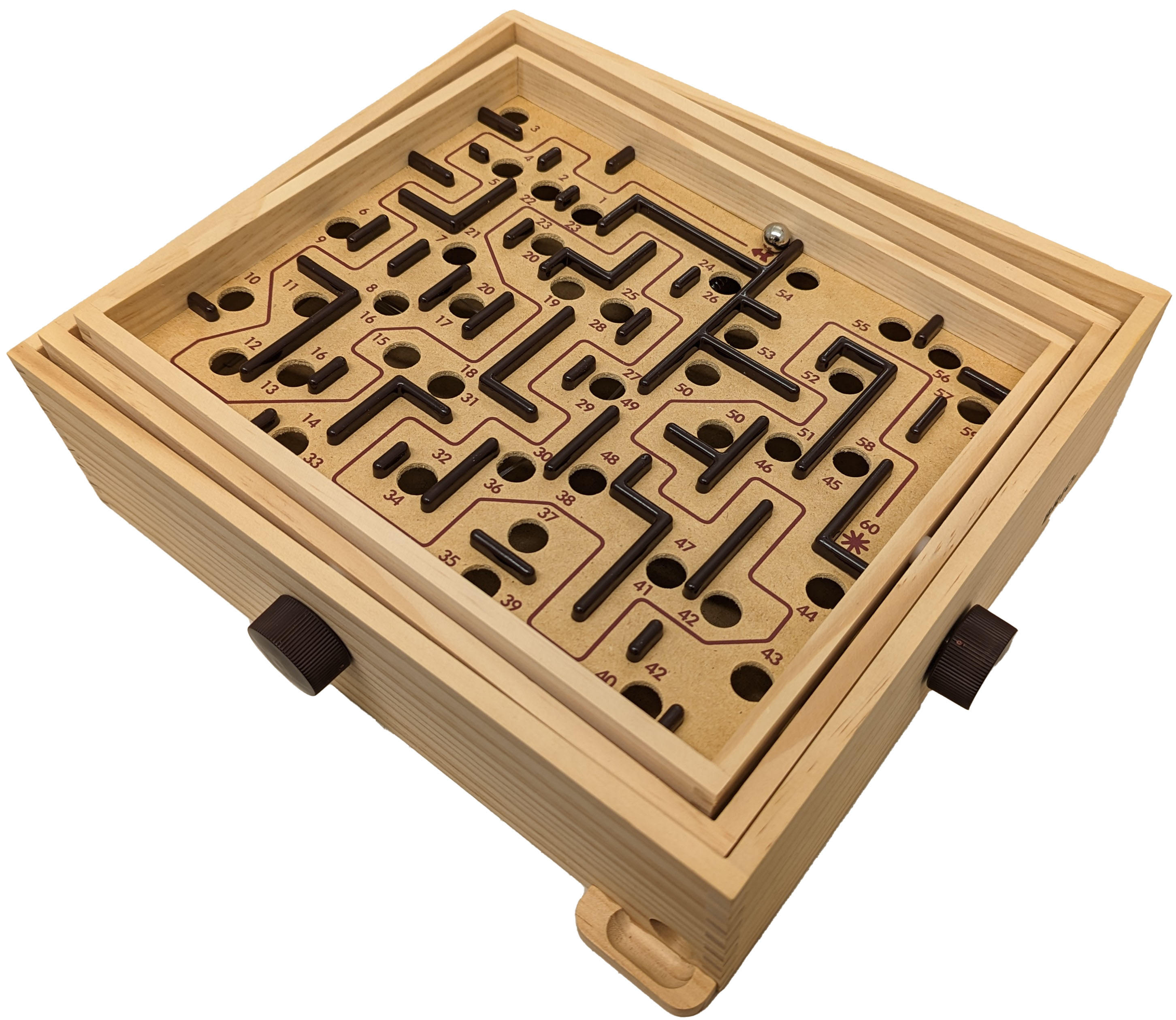}
    \caption{The labyrinth game is a marble game with the goal of steering a ball from a start to an end position while avoiding letting the ball fall down the holes. Pictured above is the BRIO Labyrinth, introduced almost 80 years ago, and with millions sold. }
    \label{fig:brio_labyrinth}
\end{figure}

\setcounter{footnote}{1} 
The labyrinth marble game has existed in various forms for many years and has been popularized in 1946 by BRIO. 
Depicted in \Cref{fig:brio_labyrinth}, the goal of the game is to maneuver a steel ball through the labyrinth without falling into one of the dozens of holes. To control the ball, the human player rotates two knobs on the side of the labyrinth to tilt the playing surface and can therefore accelerate the ball. The careful movements necessary to keep control over the ball can be very challenging for a beginner. Nevertheless, the slowly starting but rewarding learning process can make the game fairly addictive and humans can achieve remarkable results through continuous training\footnote{recordsetter.com/world-record/complete-game-labyrinth/57213}. 

The labyrinth game has not only found its usage as a game but also as a research platform for applied control theory. Multiple projects in the past have used the labyrinth to develop classical controllers for steering the ball through the labyrinth \cite{LQR_approach}. Compared to a regular ball-plate system, the labyrinth offers a few additional challenges apart from just following the path; an uneven base plate causes unexpected angle changes; pores in the wood increase the static friction; the dynamics of the ball bouncing into walls are highly nonlinear; the strings and springs that are connecting the knobs with the labyrinth plate cause slippage and a delay which again results in nonlinear and partially random behavior. These dynamics are hard to model and, therefore, challenging for classical controllers. 

With the increasing speed of hardware available and development in the area of optimization, model predictive control (MPC) has become more powerful in recent years. 
Especially in robotics, MPC is widely used \cite{MPCreview}. With the development of applicable nonlinear solvers, the area of MPC covers a whole new range of engineering possibilities \cite{FORCESNLP}. While linear MPC is already found to perform better on tracking tasks on ball-plate-systems \cite{MPC-ball-plate}, the walls of the labyrinth introduce a highly nonlinear environment which is hard to incorporate into linear optimization problems. The goal of this work is to investigate the possibilities that nonlinear and non-convex optimization brings to solve the labyrinth game and comparable robotic systems.

Given the aforementioned nonlinearities and non-convexity for the given problem, we propose to split the control problem into two controllers. The first, high-level model predictive controller continually optimizes a trajectory of the ball through the labyrinth that aims to (1) avoid any contact with the walls of the labyrinth, and (2) avoid falling down any of the holes. The complexity and non-convexity of the given optimization problem result in a low control bandwidth. Therefore, a second low-level model predictive controller is introduced that aims to follow the trajectory given by the high-level controller in real-time using a simplified model. Further, a map of the labyrinth surface irregularities is learned. 
The proposed control architecture successfully navigates the ball from start to finish with a 28\% success rate, and, on average, completes 64.4\% of the labyrinth.

\subsection{Related Work}

In recent years, learning-based approaches have become more common to solve the labyrinth \cite{bi2023sample, LWPR, RL-testbed, circle_maze}. Most notably, the authors of \cite{bi2023sample} applied model-based RL to successfully solve the labyrinth game, achieving promising results by beating the best human player.

There are, however, only a few classical control approaches published that directly try to solve the labyrinth. Andersen et al. \cite{andersen1993real}  solve the original BRIO labyrinth with a frequency domain-based controller. They claim a success rate of 75\%. However, no timings and detailed results are available. Frid et al. \cite{LQR_approach} propose a gain-scheduled LQR controller that achieves a ~78 \% success rate of finishing the labyrinth. Nevertheless, they use a simplified version of the labyrinth. Frid et al. changed the base-plate to a more straight and flat board to avoid disturbances. Additionally, they changed the string and spring-driven tilting mechanism of the labyrinth with a stiff connection to the servo. These are two of the main difficulties during control and therefore a comparison is not possible. 

More research is available related to the control of a classical ball-plate-system. Fan et al. \cite{FAN2004297} propose a hierarchical fuzzy control scheme to track a given path. Bang et al. \cite{MPC-ball-plate} and Zarzycki et al. \cite{MPC-ball-plate2} both propose model-predictive controllers. They are based on the linearized dynamics of the ball on the angled plate. Bang et al. highlights the advancements that the recent progress in MPC development brings. While classical common strategies for controlling a ball-plate system such as LQR, sliding mode control, fuzzy control, and PID perform well on stabilizing the ball, MPC performs especially better for tracking problems. To the knowledge of the authors, there have been no publications on path-tracking problems with MPC for balls that include obstacle avoidance. 


\subsection{Outline}
The hardware setup is described in \Cref{hardware}. In \Cref{method} we describe our method, starting with the derivation of the state-space model (\ref{state-space-model}) and the state estimation (\ref{state-estimation}), followed by the the detailed description of our MPC strategy. We finish with the performance results in \Cref{results} and conclude the paper in \Cref{conclusion}.
\section{Hardware}
\label{hardware}

\begin{figure}
    \centering
    \includegraphics[width=7cm]{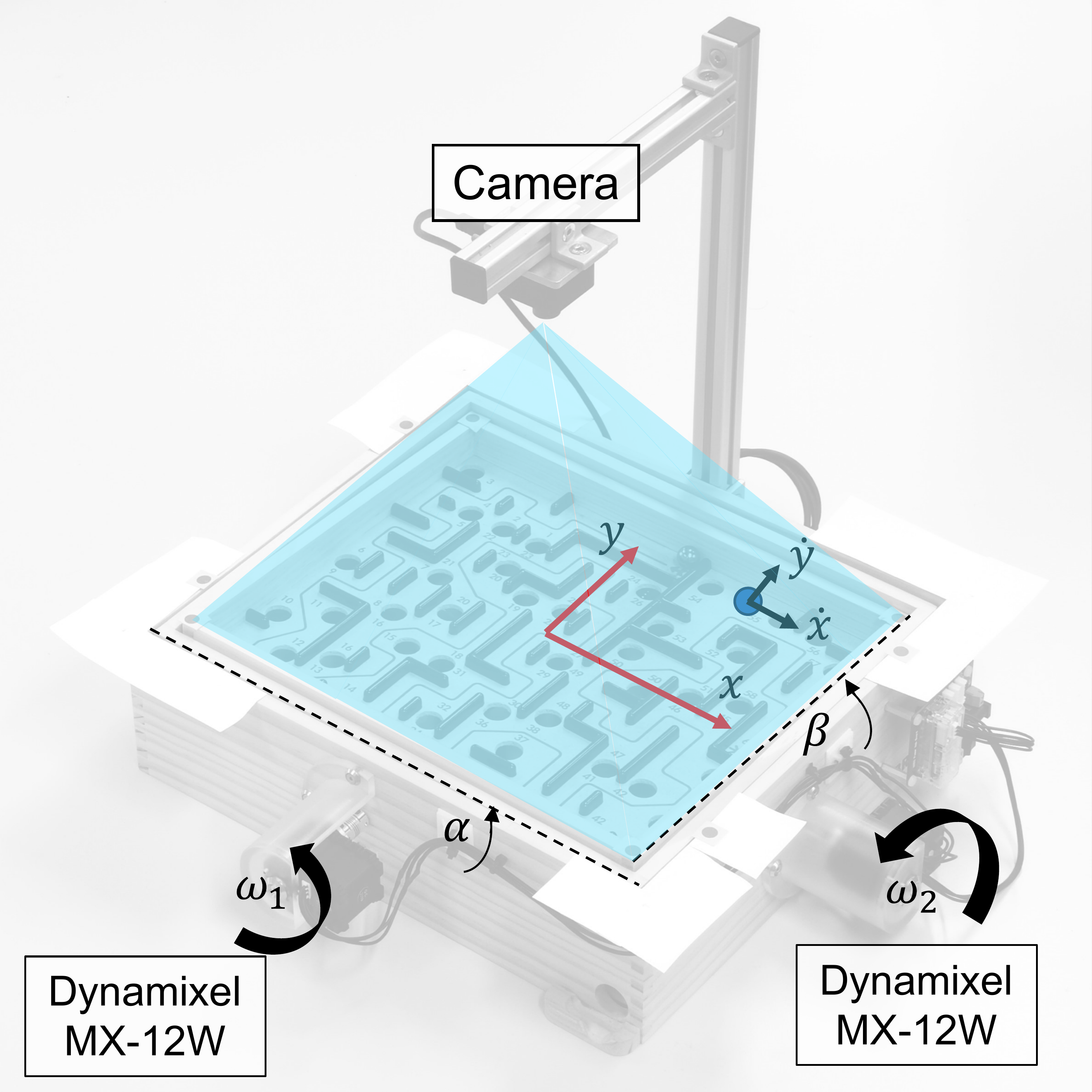}
    \caption{Hardware setup of the robotic system.}
    \label{fig:hardware-setup}
\end{figure}

The hardware of the robotic system is shown in \Cref{fig:hardware-setup}. The labyrinth itself consists of a wooden frame around the base-plate. The base-plate is gimbal-mounted on two axes, so that both angles of the labyrinth can be rotated independently by turning the axes. We attach two velocity-controlled Dynamixel MX-12W motors to the two axes. This serves as the input to the system. 
A camera is mounted on top, pointing down on the labyrinth, and provides RGB images with 55 Hz at a resolution of 1920x1200 px. The camera and the motors are connected to a desktop workstation with an AMD Threadripper Pro 5955WX CPU. 
ROS2 \cite{macenski2022robot} is used for process handling. 

\section{Method}
\label{method}
We use a camera-based state estimator and a linearized state-space model of the ball-plate system. Based on this, our method is using a two-layered MPC approach. A low-frequency high-level solver is responsible for pseudo-global optimal path-planning and a high-frequency low-level solver is responsible for path-tracking and actuation. Additionally, we propose two modules for disturbance compensation.

\subsection{State-Space Model}
\label{state-space-model}

\begin{figure}
    \centering
    \includegraphics[height=1.8cm]{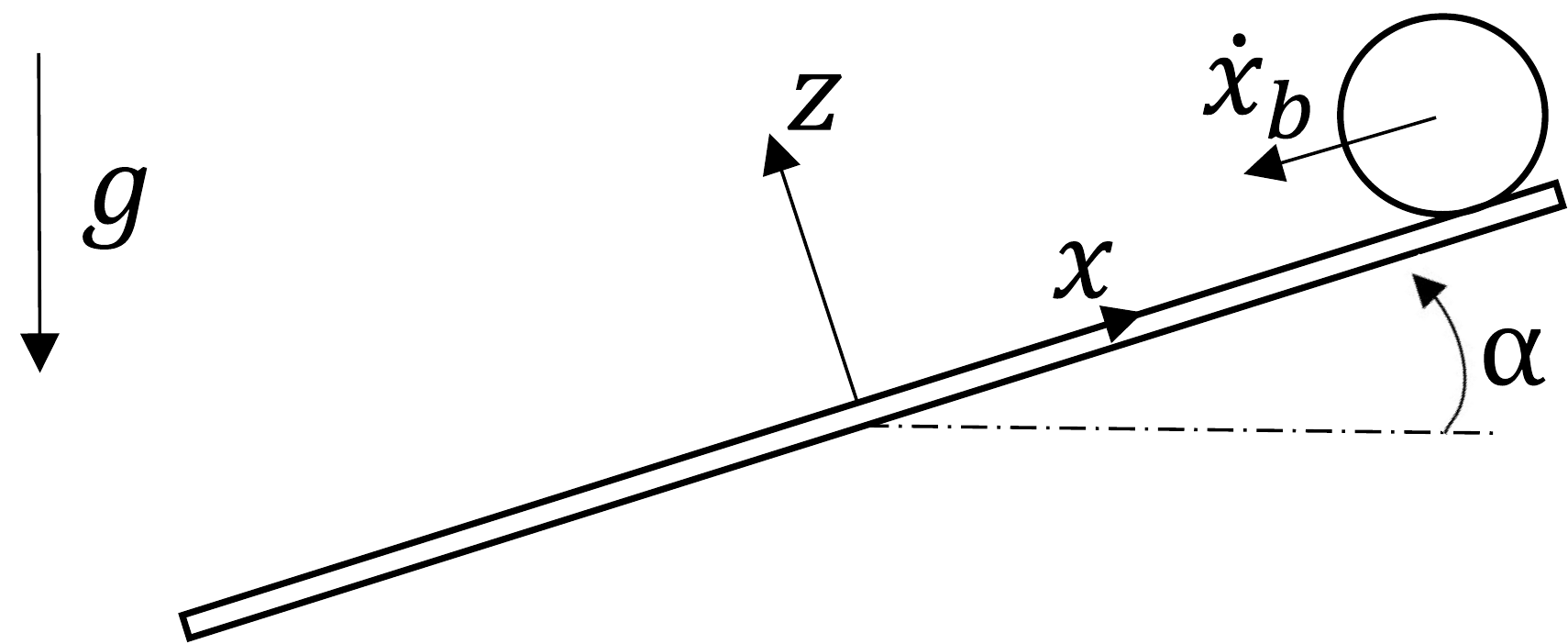}
    \caption{Schematic representation of a ball-plate-system.}
    \label{fig:ball-plate-system}
\end{figure}

Considering only the ball-plate system without obstacles or friction, the dynamics for a solid ball are given by \cite{FAN2004297} as
\begin{align}
    m_{ball}\left(\frac{7}{5} \Ddot{x}_b - \left(x_b\dot{\alpha}^2 + y_b\dot{\alpha}\dot{\beta}\right) + g\sin{\alpha}\right)=0,\\
    m_{ball}\left(\frac{7}{5} \Ddot{y}_b - \left(y_b\dot{\beta}^2 + x_b\dot{\alpha}\dot{\beta}\right) + g\sin{\beta}\right)=0,
    \label{eq:nonlinear_model}
\end{align}
where $m_b$ is the mass of the ball, $(x_b, y_b)$ is the ball's position within the plate frame, and $(\alpha, \beta)$ are the plate inclination angles (see \Cref{fig:ball-plate-system}). 
Assuming small angles and angle velocities, we have $\alpha, \beta \approx 0 \Rightarrow \sin{\alpha} \approx \alpha, \sin{\beta} \approx \beta$ and $\dot{\alpha}, \dot{\beta} \ll 1 \Rightarrow \dot{\alpha}^2, \dot{\beta}^2, \dot{\alpha}\dot{\beta} \approx 0$ and obtain
\begin{equation}
    \Ddot{x}_b \approx -\frac{5}{7}g\alpha, \quad \Ddot{y}_b \approx -\frac{5}{7}g\beta .
    \label{eq:linear_dynamics}
\end{equation}

\begin{figure}
    \centering
    \includegraphics[height=2.1cm]{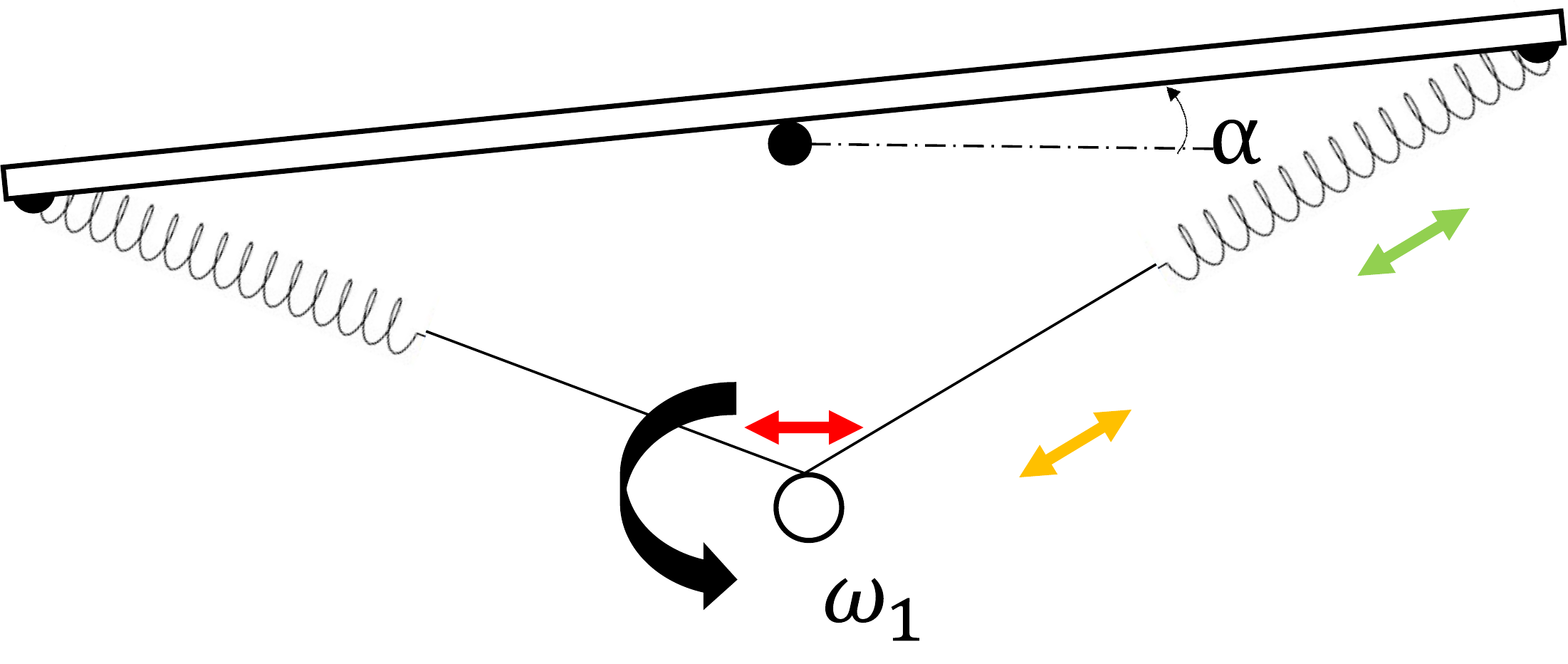}
    \caption{Schematic representation of the plate-tilting-mechanism}
    \label{fig:plat-tilting-mechanism}
\end{figure}

The plate-tilting mechanism is shown in \Cref{fig:plat-tilting-mechanism}. The plate is pulled by two springs, which are connected by a string that is wrapped around the motor's axis. The motor with rotational velocity $\omega_1$ steers $\alpha$ (and $\omega_2$ steers $\beta$ accordingly). 
The relation of $\alpha$ to $\omega_1$ is nonlinear. Slippage of the string around the axis and situations where one side of the string is not under tension make the connection hard to model accurately. Assuming only small changes of $\alpha$ and $\beta$, we use an approximated linear relation
\begin{equation}
    \dot{\alpha} \approx k_1\omega_1, \quad \dot{\beta} \approx k_2\omega_2 .
    \label{eq:tilt}
\end{equation}

Finally, the state and inputs are defined as
\begin{align}
    \hat{x} &= \begin{bmatrix}x_b, \dot{x_b}, y_b, \dot{y_b}, \alpha, \beta\end{bmatrix}^T,\\
    \hat{u} &= [\omega_1, \omega_2]^T.
\end{align}
Then, assuming a zero-order hold on the input, the time-discretized state-space model with the discrete time step $T_S$ is given by
\begin{gather}
    A = \small\setlength\arraycolsep{2pt}
    \begin{bmatrix}
        1 & T_S & 0 & 0 & 0 & 0 \\
        0 & 1 & 0 & 0 & -\frac{5}{7}gT_S & 0 \\
        0 & 0 & 1 & T_S & 0 & 0 \\
        0 & 0 & 0 & 1 & 0 & -\frac{5}{7}gT_S \\
        0 & 0 & 0 & 0 & 1 & 0 \\
        0 & 0 & 0 & 0 & 0 & 1 \\
    \end{bmatrix},
    B = \small\setlength\arraycolsep{2pt}
    \begin{bmatrix}
        0 & 0 \\
        0 & 0 \\
        0 & 0 \\
        0 & 0 \\
        k_1T_S & 0 \\
        0 & k_2T_S \\
    \end{bmatrix},\\
    \hat{x}_{k+1} = A\hat{x}_k + B\hat{u}_k, \quad \forall k \geq 0 .
    \label{eq:state-space-model}
\end{gather}

\begin{figure}
    \centering
    \includegraphics[height=3.5cm]{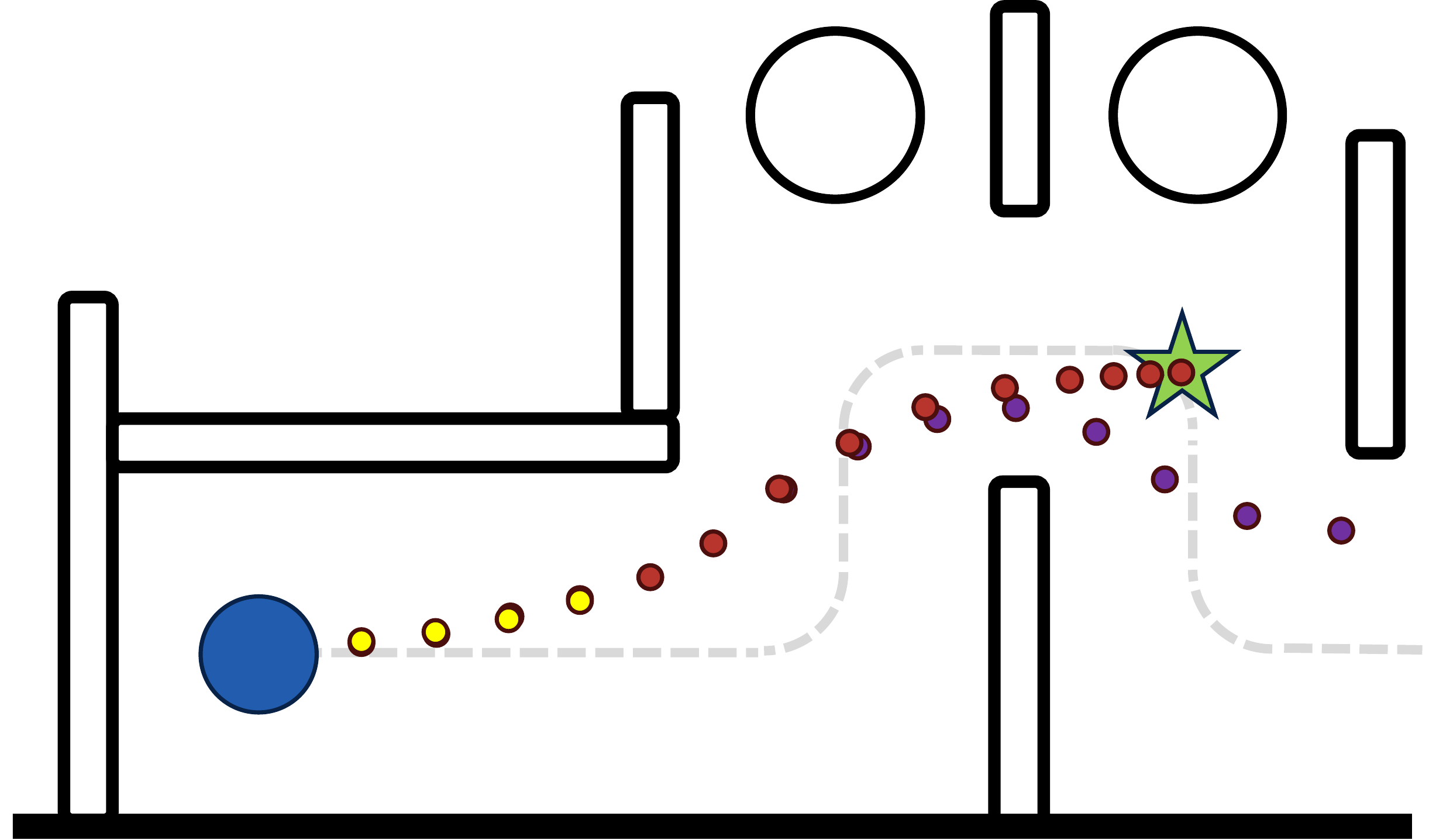}
    \caption{Concept of the non-linear MPC approach. Blue: Ball. Green star: Current goal. Red: High-level path. Yellow: Low-level path. Purple: Global optimal path around obstacles.}
    \label{fig:nonlinear-concept}
\end{figure}

\subsection{State Estimation}
\label{state-estimation}

With color- and contour-based image masking, the blue ball and markers on the corners of the labyrinth are tracked in the current camera image, and the ball's position $(x,y)$, velocity $(\dot{x},\dot{y})$ and the labyrinth's base-plate angles $(\alpha,\beta)$ are estimated using a Kalman Filter based on the linear state-space model. A detailed description of the state estimation can be found in \cite{bi2023sample}.

\subsection{Model Predictive Control}
\label{model-predictive-control}
A typical situation from the labyrinth is shown in \Cref{fig:nonlinear-concept}. The task of our high-level path-planning MPC (HL-solver) is to find or approximate the optimal path around the obstacles (purple). Based on the current ball position, we choose the third next corner on the underlying perpendicular path (grey) as the current goal (green star) for the HL-solver and update it in a receding horizon scheme. A more distant goal could result in more optimal paths but makes the problem harder to solve. The corner locations of the underlying perpendicular path are given to the system. We incorporate the ball-plate system dynamics and set the obstacles as constraints. The solution that the HL-solver finds is a locally optimal path (red), which approximates the globally optimal path. To track the path, we propose a low-level MPC (LL-solver) that computes a control input for each new state estimate at 55Hz. It incorporates the ball-plate-system dynamics but is not constrained by the obstacles, which reduces the solving time drastically. Splitting the system in a slow HL-solver and a fast LL-solver allows us to apply the optimal solutions of computationally heavy MPC in a high-frequency real-time environment. An overview can be seen in \Cref{tab:solver_comparison}. The unit of length in the problem formulations is m, angles in rad, time in s and the weights are unitless.


\renewcommand\tabularxcolumn[1]{m{#1}} 
\newcolumntype{Y}{>{\centering\arraybackslash}X} 

\begin{table}
\centering
\caption{Comparison of High-level and Low-level solvers}
\begin{tabularx}{8cm}{@{} >{\bfseries}l | Y | Y @{}}
\toprule
& \makecell{\textbf{High-level solver}} & \makecell{\textbf{Low-level solver}} \\
\midrule
\makecell{Horizon length} & \makecell{$N = 100$} & \makecell{$N = 18$} \\
\hline
\makecell{Time step} & \makecell{$T_s = 30$ ms} & \makecell{$T_s = 30$ ms} \\
\hline
\makecell{Solving time} & \makecell{$\approx 100$ ms} & \makecell{$\approx 5$ ms} \\
\hline
\makecell{Control frequency} & \makecell{$\approx 10$ Hz} & \makecell{$55$ Hz} \\
\hline
\makecell{Cost-function} & Ball to goal distance, \( u, \dot{x}_{b} \) & Difference to HL-path \( (x_{b}, \dot{x}_{b}, u) \), holes, walls \\
\hline
\makecell{Inter-stage equality} & \multicolumn{2}{c}{\( x_{k+1} = Ax_{k} + Bu_{k} \)} \\
\hline
\makecell{Inequality constraints} & Holes, Walls & \multicolumn{1}{c}{---} \\
\hline
\makecell{Terminal Constraint} & Current goal & \( 15^{th} \) point on HL-path from current position \\
\hline
\makecell{Boundaries} & \multicolumn{2}{c}{Max. input, labyrinth boundaries} \\
\bottomrule
\end{tabularx}
\label{tab:solver_comparison}
\end{table}

\subsubsection{High-Level Solver}

To solve the nonlinear non-convex optimization problem we use the Primal-Dual Interior-Point method provided by Embotech's ForcesPRO \cite{FORCESNLP}. The optimization problem is formulated as
\begin{align}
    &\underset{\hat{x}_k, \hat{u}_k}{\text{minimize}} \quad \sum_{k=1}^{N} (\hat{x}_k - \hat{x}_r)^T Q (\hat{x}_k - \hat{x}_r) + \hat{u}_k^T R \hat{u}_k \\
    \label{eq:interstage-high-level}
    &\text{subject to} \quad \hat{x}_{k+1} = A\hat{x}_k + B\hat{u}_k\\
    &\phantom{\text{subject to} \quad} \begin{bmatrix}
        -x_{frame}\\
        -\infty\\
        -y_{frame}\\
        -\infty\\
        -\alpha_{max}\\
        -\beta_{max}\\
    \end{bmatrix} \leq \hat{x}_k \leq \begin{bmatrix}
        x_{frame}\\
        \infty\\
        y_{frame}\\
        \infty\\
        \alpha_{max}\\
        \beta_{max}\\
    \end{bmatrix}\\
    &\phantom{\text{subject to} \quad} -\omega_{max}\leq \hat{u}_k[i] \leq \omega_{max},\ i=1,2\\
    &\phantom{\text{subject to} \quad} \hat{x}_0 = \hat{x}_{est}\\
    &\phantom{\text{subject to} \quad} \hat{x}_N[1] = x_{goal}, \quad \hat{x}_N[3] = y_{goal}\\
    &\phantom{\text{subject to} \quad} \hat{h}_{lower} \leq h(\hat{x}_k, \hat{u}_k, \hat{p}) \leq \hat{h}_{upper}
    \label{eq:non-linear-inequalities}
\end{align}

The objective function's focus lies on reaching the goal while ensuring smooth behavior through small velocity and input weights
\begin{equation}
    Q = \diag(10, 1, 10, 1, 0, 0), \quad R = \diag(0.1, 0.1)
\end{equation}

The reference state accordingly is
\begin{equation}
    \hat{x}_r = [x_{goal}, 0, y_{goal}, 0, 0, 0]^T .
\end{equation}

\begin{figure}
    \centering
    \includegraphics[height= 3.6cm]{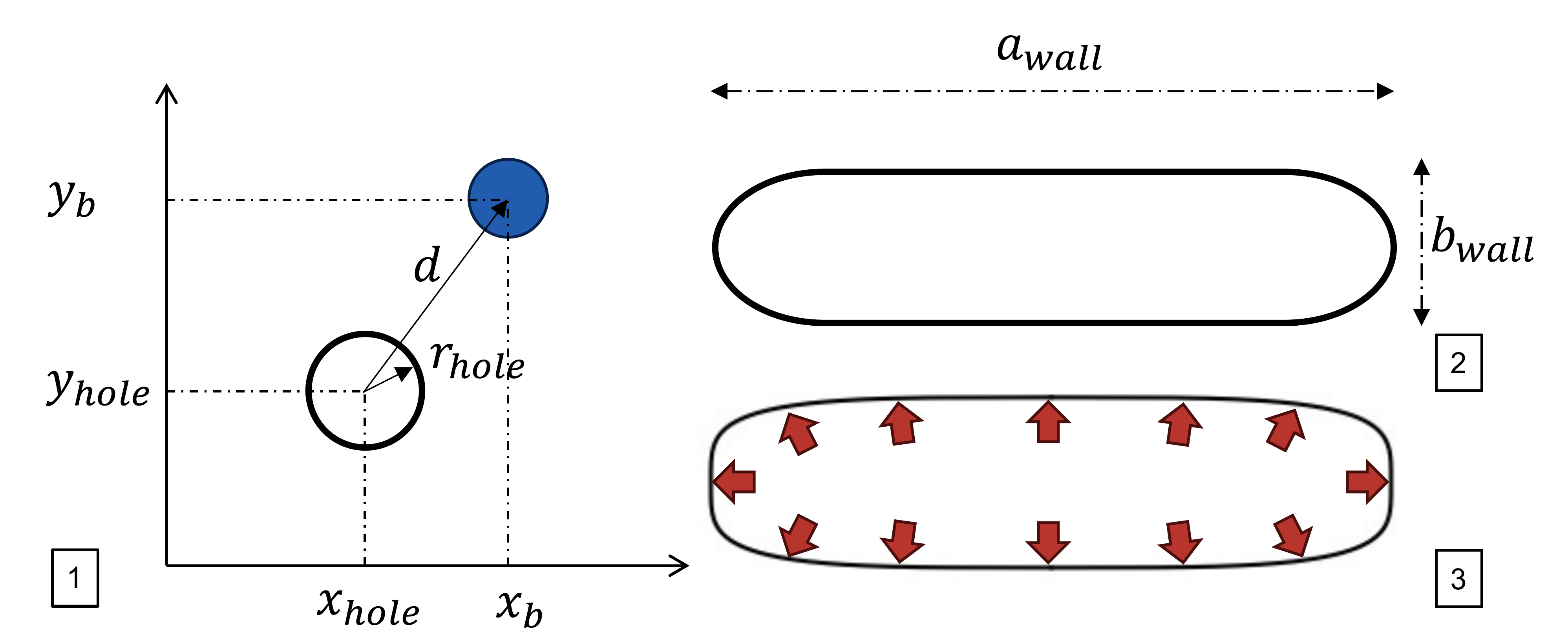}
    \caption{\scalebox{0.7}{\fbox{1}} - Ball next to hole. \scalebox{0.7}{\fbox{2}} - Wall shape. \scalebox{0.7}{\fbox{3}} - Superellipse approximation of the wall with exponent $f=4$. The red arrows show the gradient of \Cref{eq:superellipse}}.
    \label{fig:obstacle-constraints}
\end{figure}

We constrain the state and input using the labyrinth frame ($x_{frame},y_{frame}$) and reduced values of the maximum possible angles ($\alpha_{max},\beta_{max}$) and servo velocity $\omega_{max}$. The initial and final constraints are set to the current state estimate $\hat{x}_{est}$ and the current goal ($x_{goal}, y_{goal}$) (3rd next corner). The inequality constraints are formulated as a stack of nonlinear functions $h(\cdot)$ with fixed upper and lower bounds. They incorporate the obstacles, i.e., the walls and holes. The distance of the ball center to the hole center (see \Cref{fig:obstacle-constraints}) is
\begin{equation}
    d = \sqrt{(x_b-x_{hole})^2 + (y_b-y_{hole})^2} .
\end{equation}
In order that the ball does not fall into the hole we constrain $d \geq r_{hole}$. We reformulate to the form of Equation~(\ref{eq:non-linear-inequalities})
\begin{equation}
    r_{hole}^2 \leq (x_b-x_{hole})^2 + (y_b-y_{hole})^2 \leq \infty .
\end{equation}

To approximate the shape of the walls, we use a superellipse. The contour of an origin-centered superellipse is given by
\begin{equation}
    \left| \frac{x}{a_{wall}} \right|^{f} + \left| \frac{y}{b_{wall}} \right|^{f} = 1,
\end{equation}
where $a_{wall}$ and $b_{wall}$ are the length and width of the wall, respectively. 
To account for the ball size, we formulate our constraint, so that only ball positions with a distance higher than $r_{ball}$ to the wall are allowed. An exponent $f=4$ approximates the wall shape well and has the gradient pointing slightly sideways (red arrows in \Cref{fig:obstacle-constraints}) to help the solver find solutions around the obstacle. This approach leverages the fact that all walls on the labyrinth are horizontally or vertically positioned. Then, the constraint of not hitting the walls is formulated as
\begin{equation}
\label{eq:superellipse}
    1 \leq \left( \frac{x_b - x_{\text{wall}}}{a_{\text{wall}} + r_{\text{ball}}} \right)^4 + \left( \frac{y_b - y_{\text{wall}}}{b_{\text{wall}} + r_{\text{ball}}} \right)^4 \leq \infty,
\end{equation}
where $(x_{wall}, y_{wall})$ denotes the center of the wall.

To reduce computation time, a look-up table is used to only choose the closest 5 holes and 10 walls to the ball to incorporate in the optimization problem. Obstacle positions and size are given to $h(\cdot)$ as run-time parameters $\hat{p}$. Especially for non-convex problems, the initial guess for the solver is crucial for convergence to the right solution. According to choosing the goal, we use the second next corner of the labyrinth ($x_{co}, y_{co}$) as the initial guess
\begin{equation}
    \hat{x}_{k,init} = [x_{co},0,y_{co},0,0,0]^T 
 \quad \forall k=1,2,..,N .
\end{equation}

The HL-solver yields a feasible 100-step path including ball position, velocity, angles, and system inputs. In case the solver doesn't find a solution for 200 ms, we terminate and restart the solver from the new state. Until the next solution is found, we use the last feasible HL-solution.

\subsubsection{Low-Level Solver}

We formulate the optimization problem as
\begin{align}
    &\underset{\hat{x}_k, \hat{u}_k}{\text{minimize}} \quad \sum_{k=1}^{N} \biggl( (\hat{x}_k - \hat{x}_{r,k})^T Q (\hat{x}_k - \hat{x}_{r,k}) \\
    &\quad + (\hat{u}_k - \hat{u}_{r,k})^T R (\hat{u}_k - \hat{u}_{r,k}) + w_{obs}c_{obs}(\hat{x}_k) \biggl) \nonumber\\
    \label{eq:interstage-low-level}
    &\text{subject to} \quad \hat{x}_{k+1} = A\hat{x}_k + B\hat{u}_k\\
    &\phantom{\text{subject to} \quad} \begin{bmatrix}
        -\infty\\
        -\infty\\
        -\infty\\
        -\infty\\
        -\alpha_{max}\\
        -\beta_{max}\\
    \end{bmatrix} \leq \hat{x}_k \leq \begin{bmatrix}
        \infty\\
        \infty\\
        \infty\\
        \infty\\
        \alpha_{max}\\
        \beta_{max}\\
    \end{bmatrix}\\
    &\phantom{\text{subject to} \quad} -\omega_{max} \leq \hat{u}_k \leq \omega_{max}\\
    &\phantom{\text{subject to} \quad} \hat{x}_0 = \hat{x}_{est}\\
    &\phantom{\text{subject to} \quad} \hat{x}_N[1] = x_{HL,15}, \quad \hat{x}_N[3] = y_{HL,15}
\end{align}

\begin{figure}
    \centering
    \includegraphics[height=1.7cm]{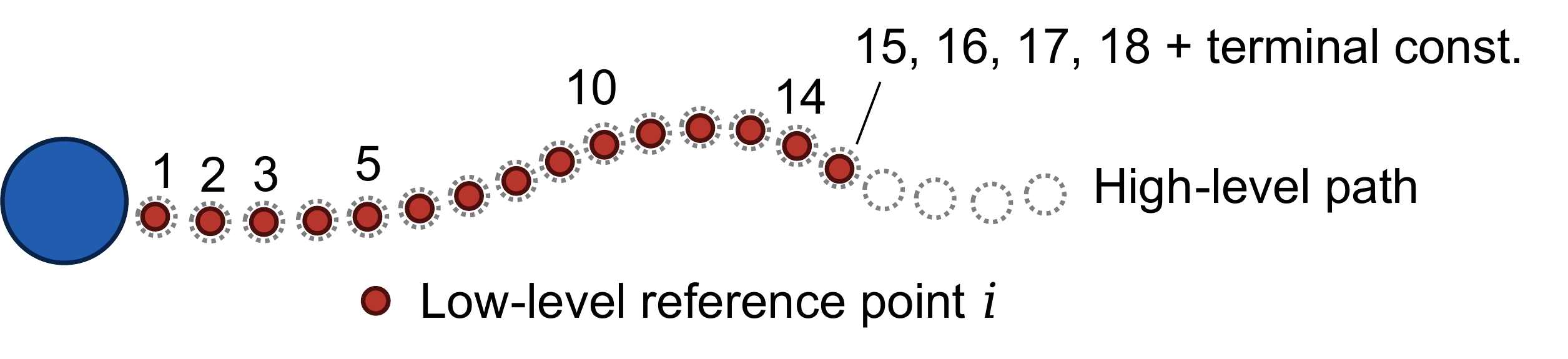}
    \caption{Depiction of which points from the high-level path get used as a reference for the low-level path.}
    \label{fig:lowlevel-reference}
\end{figure}

As a reference for the state and input, we use the 15 next points on the high-level path starting from the current ball position (\Cref{fig:lowlevel-reference}). We use the 15th point as the terminal constraint and as the last four low-level references as well to have a buffer and ensure solvability.

The reference vectors are given by
\begin{equation}
    \hat{x}_{r,k} = \begin{bmatrix}
    x_{HL,i} \\
    \dot{x}_{HL,i} \\
    y_{HL,i} \\
    \dot{y}_{HL,i} \\
    0 \\
    0 \\
\end{bmatrix}, \quad
\hat{u}_{r,k} = \begin{bmatrix}
    \omega_{1,HL,i} \\
    \omega_{2,HL,i} \\
\end{bmatrix},
\label{eq:reference-state}
\end{equation}
\begin{equation}
    i=\begin{cases}
        k & \text{if } k \leq 15\\
        15 & \text{if } 16\leq k \leq 18
    \end{cases}
    \quad \forall k = 1,2, .. ,N, \nonumber
\end{equation}
where $_{HL,i}$ denotes the corresponding $i$-th value on the high-level path starting from the current ball position. The weights focus on positional correctness
\begin{equation}
\begin{split}
    &Q = \diag(10, 0.1, 10, 0.1, 0, 0), \quad R = \diag(0.3, 0.3),\\
    &w_{obs} = 0.01.
\end{split}
\end{equation}
Additionally, we introduce a cost term for the $J = 15$ nearest obstacles to avoid short-cuts of the low-level path through obstacles
\begin{equation}
    c_{obs} = \sum^{J}_{j=1} c_{obs,j}.
\end{equation}

We use the center locations of the holes and the two end points of each wall as the obstacle locations $(x_{obs,j}, y_{obs,j})$. The distance of the ball to each obstacle $j$ is
\begin{equation}
    d_{obs,j} = \sqrt{(x_b - x_{obs,j})^2 + (y_b - y_{obs,j})^2} .
\end{equation}

We feed it through an approximated differentiable ReLu function to only penalize the ball if it gets closer than $d_{max} = 0.0125$  to the obstacle center
\begin{equation}
    c_{obs,j} = \log\left(1+e^{10000\cdot(-d_{obs,j} + d_{max})}\right) .
\end{equation}

We set the initial guess for the solver according to the current state estimate
\begin{equation}
    \hat{x}_{k,init} = [x_{b},0,y_{b},0,0,0]^T 
 \quad \forall k=1,2,..,N .
\end{equation}
Solving the optimization problem at every 55Hz control cycle yields the next input to the system.

\subsubsection{Disturbance Compensator}
\label{disturbance-compensator}

Addressing the unevenness of the labyrinth's base plate, we introduce a Luenberger state estimator \cite{Luenberger1099826} for disturbance compensation. We  calculate the state prediction for the current state $\Tilde{x}_{k+1}$ based on the last state $\hat{x}_k$, input $\hat{u}_k$ and disturbance estimate $\Tilde{d}_k \in \mathbb{R}^2$
\begin{equation}
    \Tilde{x}_{k+1} = A\hat{x}_k + B\hat{u}_k + B_d \Tilde{d}_k .
    \label{eq:disturbance_comp}
\end{equation}

The adjusted disturbance estimate $\Tilde{d}_{k+1}$ is received from the difference between the actual state $\hat{x}_k$ and the predicted state $\Tilde{x}_k$
\begin{equation}
    \Tilde{d}_{k+1} = \Tilde{d}_k + L(\hat{x}_k - \Tilde{x}_k).
    \label{eq:disturbance-estimate}
\end{equation}

We set
\begin{equation}
    B_d = \begin{bmatrix}
        0 & 0 \\
        0 & 0 \\
        0 & 0 \\
        0 & 0 \\
        1 & 0 \\
        0 & 1 \\
    \end{bmatrix}, \quad
    L = \begin{bmatrix}
        0 & 0 \\
        0.04 & 0 \\
        0 & 0 \\
        0 & 0.04 \\
        0 & 0 \\
        0 & 0 \\
    \end{bmatrix}^T,
\end{equation}
where $L$ contains the tuned learning weights. \Cref{eq:disturbance_comp} replaces the inter-stage equality term in the problem formulations (\Cref{eq:interstage-high-level} and \Cref{eq:interstage-low-level}).

Since the walls are not represented in the state-space model, the disturbance estimator sees walls as disturbances as well. To avoid false changes in the disturbance estimate while hitting walls, we introduce a heuristic, which checks if the predicted ball position $\Tilde{x}_k$ is close to a wall and the correction vector $\hat{x}_k - \Tilde{x}_k$ is pointing towards that wall. In that case, we set $\Tilde{d}_{k+1} = \Tilde{d}_k$ for one control cycle.

\subsubsection{Feed-Forward Angle Map}
\label{angle-map}

\begin{figure}
    \centering
    \includegraphics[height=3.8cm]{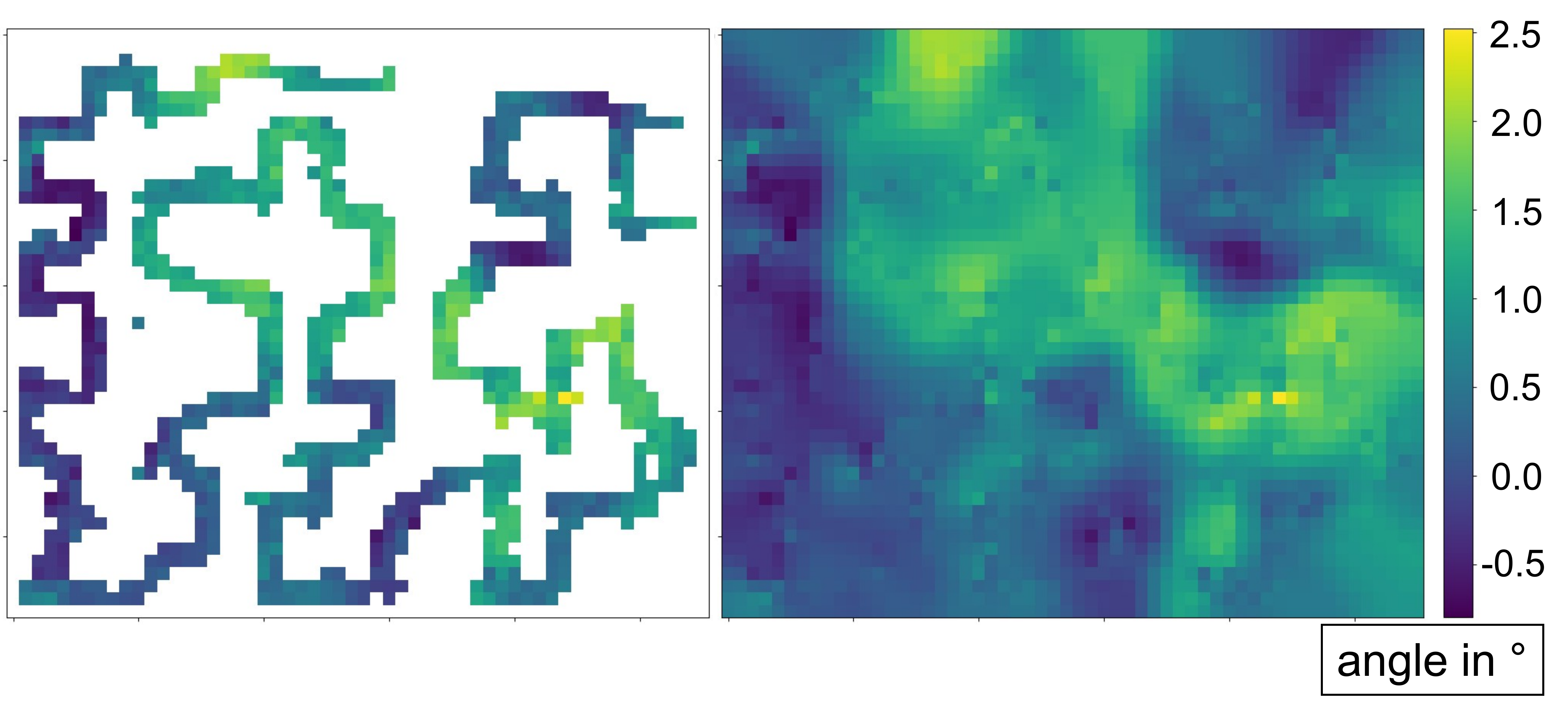}
    \caption{Left: Measured and averaged angles obtained by the disturbance compensator. Right: Interpolated final angle map (angles in $x$-direction).}
    \label{fig:angle-map}
\end{figure}

The feedback-based disturbance compensator detects disturbances with delay. A major part of the disturbances comes from the uneven base-plate. These stay the same in every run of the labyrinth and can therefore be applied in a feed-forward way. During multiple runs, backwards and forwards, we record the disturbances measured by the disturbance compensator to create an angle map. We add the measured angles to the state estimate in the next run and record an angle map again to obtain finer measurements. After three iterations, the recorded angles converge. We interpolate the missing regions and received two final angle maps in x- and y-direction (\Cref{fig:angle-map}).

The angle map contains enough information about the disturbances to do full runs of the labyrinth without the disturbance compensator. However, better performance is achieved by combining both the feed-forward map and the disturbance compensator.

\section{Results}
\label{results}

\begin{figure}
    \centering
    \includegraphics[height = 5.7cm]{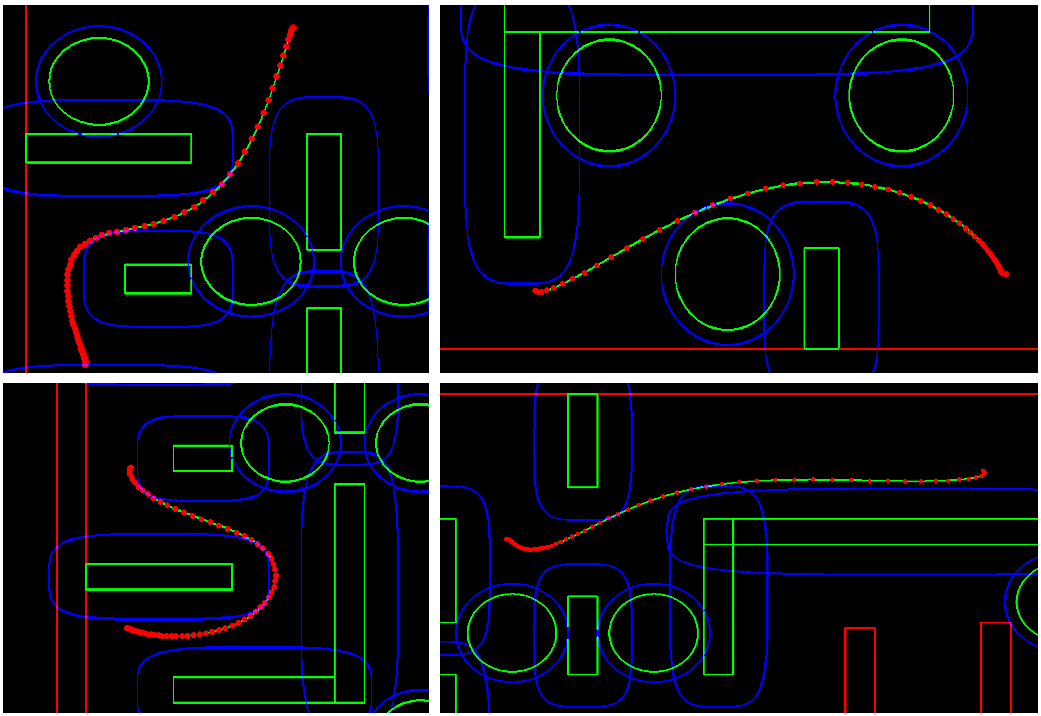}
    \caption{Various labyrinth setups. The red dotted line shows the result of the high-level MPC from the ball position to the 3rd next corner. Blue lines show the obstacle constraints, which are larger due to the ball’s radius.}
    \label{fig:high-level-paths}
\end{figure}

Trajectories generated by the high-level solver can be seen in \Cref{fig:high-level-paths}. The high-level solver plans paths reliably and the low-level solver is able to follow the path. Full runs of the labyrinth can be achieved regularly. As expected, difficulties occur mostly if the ball hits the wall, since these dynamics are not represented in the model. The performance of a full run of the labyrinth can be seen in \href{https://youtu.be/2LkszVNbXv8}{this video}\footnote{youtu.be/2LkszVNbXv8}.

\begin{figure}
    \centering
    \includegraphics[width=7cm]{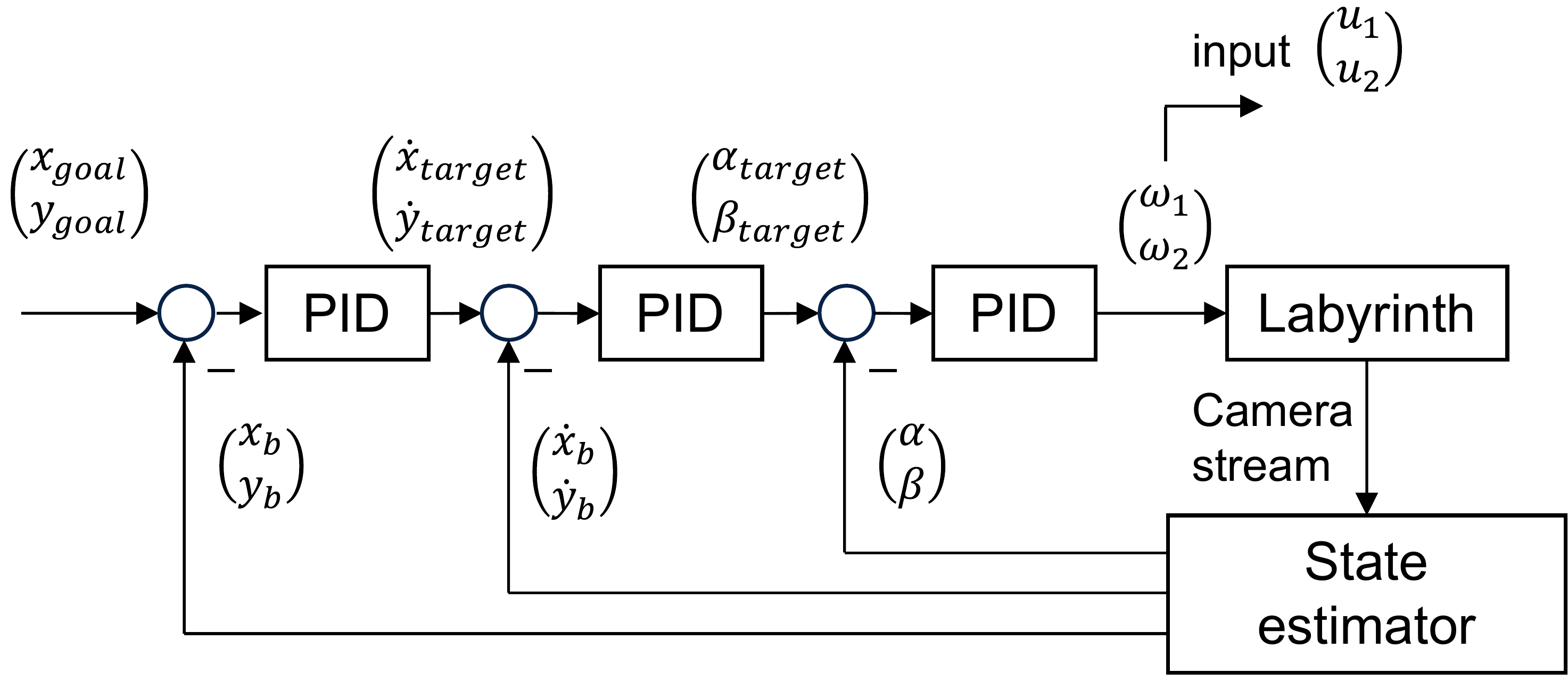}
    \caption{Cascaded PID block diagram.}
    \label{fig:PID-diagram}
\end{figure}

For comparison, we introduce a cascaded PID controller and a linear MPC contoller. Both controllers follow a trajectory of given way-points. Once the ball comes close to a way-point (7mm), the next way-point is chosen. The PID's cascaded structure is shown in \Cref{fig:PID-diagram}. The linear MPC uses the same state-space model from \Cref{eq:state-space-model} as its interstage-equality constraint. Its objective function minimizes the distance to the next way-point and penalizes ball velocity and input with small weights. Information about the obstacles is only incorporated in a way that one way point is always set so that the next one is in line of sight. The disturbance compensator (\Cref{disturbance-compensator}) and the feed-forward angle map (\Cref{angle-map}) are applied as well.

\begin{figure}
    \centering
    \includegraphics[width=8.0cm]{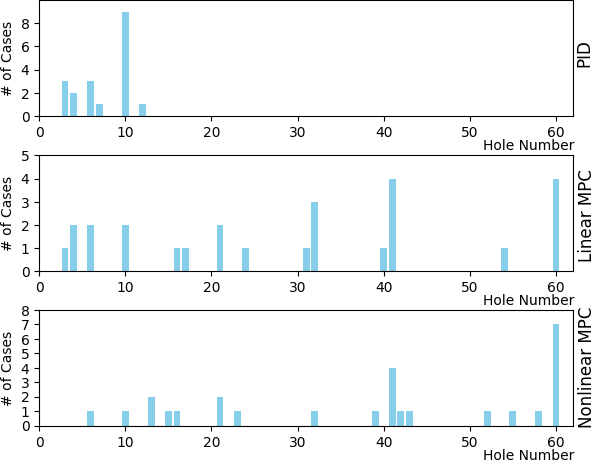}
    \caption{Comparison of how far the ball got during multiple test runs with the three different controllers.}
    \label{fig:comparison-diagram}
\end{figure}

We perform 25 runs for each controller and measure the achieved distance by means of noting the hole number where the ball fell in, and record the completion time for successful runs. These results are summarized in \Cref{tab:comparison} and depicted in \Cref{fig:comparison-diagram}. 
As can be seen, both MPC approaches are able to do full runs on the labyrinth. Nevertheless, the nonlinear, non-convex approach incorporating the obstacles in its planning achieves a higher success rate with lower completion times. The PID shows the correct behavior but is not able to complete the full labyrinth. 

\begin{table}
\centering
\caption{Comparison of different controllers}
\begin{tabular}{lccc}
\toprule
                  & \textbf{PID} & \textbf{linear MPC} & \textbf{nonlinear MPC} \\
\midrule
Full completion   & 0.0 \%            & 12.0 \%           & 28.0 \%           \\
Average distance  & 12.6 \%           & 46.3 \%           & 64.4 \%           \\
Average time      & ---               & 81.0 s            & 58.0 s            \\
\bottomrule
\end{tabular}
\label{tab:comparison}
\end{table}

\section{Conclusion}
\label{conclusion}
We presented an approach of splitting a computation-heavy MPC problem for a real-world labyrinth game into two sub-problems, to leverage the benefits of formulating a nonlinear non-convex optimization problem and still applying it to a high-frequency real-time system. We introduced a method to incorporate the differently shaped obstacles in the optimization problem and obtain a competitive controller that can deal with a variety of disturbances. The resulting controller is able to solve the labyrinth without any mechanical modifications. 
We envision that our method can be applied to similar control problems e.g. autonomous race cars on an obstacle course \cite{Zeilinger}.
In future work, we envision increasing the overall robustness of the controller in two ways. First, a more accurate model can be obtained by modeling the dynamics of the plate tilting mechanism including the spring. Second, modeling the collision of the ball with the walls could lead to more robust strategies where the collision is exploited to redirect the ball, instead of attempting to avoid the walls.

\bibliographystyle{IEEEtran}
\bibliography{references}

\begin{thebibliography}{10}
\providecommand{\url}[1]{#1}
\csname url@rmstyle\endcsname
\providecommand{\newblock}{\relax}
\providecommand{\bibinfo}[2]{#2}
\providecommand\BIBentrySTDinterwordspacing{\spaceskip=0pt\relax}
\providecommand\BIBentryALTinterwordstretchfactor{4}
\providecommand\BIBentryALTinterwordspacing{\spaceskip=\fontdimen2\font plus
\BIBentryALTinterwordstretchfactor\fontdimen3\font minus \fontdimen4\font\relax}
\providecommand\BIBforeignlanguage[2]{{%
\expandafter\ifx\csname l@#1\endcsname\relax
\typeout{** WARNING: IEEEtran.bst: No hyphenation pattern has been}%
\typeout{** loaded for the language `#1'. Using the pattern for}%
\typeout{** the default language instead.}%
\else
\language=\csname l@#1\endcsname
\fi
#2}}

\bibitem{LQR_approach}
E.~Frid and F.~Nilsson, ``Path following using gain scheduled lqr control: with applications to a labyrinth game,'' 2020.

\bibitem{MPCreview}
M.~Schwenzer, M.~Ay, T.~Bergs, and D.~Abel, ``Review on model predictive control: an engineering perspective,'' \emph{The International Journal of Advanced Manufacturing Technology}, 2021.

\bibitem{FORCESNLP}
A.~Zanelli, A.~Domahidi, J.~Jerez, and M.~Morari, ``Forces nlp: an efficient implementation of interior-point methods for multistage nonlinear nonconvex programs,'' \emph{Int. Journal of Control}, 2017.

\bibitem{MPC-ball-plate}
H.~Bang and Y.~Lee, ``Embedded model predictive control for enhancing tracking performance of a ball and plate system,'' \emph{IEEE Access}, 2019.

\bibitem{bi2023sample}
T.~Bi and R.~D'Andrea, ``Sample-efficient learning to solve a real-world labyrinth game using data-augmented model-based reinforcement learning,'' \emph{arXiv preprint arXiv:2312.09906}, 2023.

\bibitem{LWPR}
K.~Öfjäll and M.~Felsberg, ``Combining vision, machine learning and automatic control to play the labyrinth game,'' 2016.

\bibitem{RL-testbed}
J.~H. Metzen, E.~Kirchner, L.~Abdenebaoui, and F.~Kirchner, ``The brio labyrinth game-a testbed for reinforcement learning and for studies on sensorimotor learning,'' in \emph{MSRL}, 2009.

\bibitem{circle_maze}
D.~Romeres, D.~K. Jha, A.~DallaLibera, B.~Yerazunis, and D.~Nikovski, ``Semiparametrical gaussian processes learning of forward dynamical models for navigating in a circular maze,'' in \emph{IEEE ICRA}, 2019.

\bibitem{andersen1993real}
N.~A. Andersen, O.~Ravn, and A.~T. S{\o}rensen, ``Real-time vision based control of servomechanical systems,'' in \emph{Experimental Robotics II}.\hskip 1em plus 0.5em minus 0.4em\relax Springer, 1993.

\bibitem{FAN2004297}
X.~Fan, N.~Zhang, and S.~Teng, ``Trajectory planning and tracking of ball and plate system using hierarchical fuzzy control scheme,'' \emph{Fuzzy Sets and Systems}, 2004.

\bibitem{MPC-ball-plate2}
K.~Zarzycki and M.~Ławryńczuk, ``Fast real-time model predictive control for a ball-on-plate process,'' \emph{Sensors}, 2021.

\bibitem{macenski2022robot}
S.~Macenski, T.~Foote, B.~Gerkey, C.~Lalancette, and W.~Woodall, ``Robot operating system 2: Design, architecture, and uses in the wild,'' \emph{Science Robotics}, 2022.

\bibitem{Luenberger1099826}
D.~Luenberger, ``An introduction to observers,'' \emph{IEEE Transactions on Automatic Control}, 1971.

\bibitem{Zeilinger}
L.~Numerow, A.~Zanelli, A.~Carron, and M.~Zeilinger, ``Inherently robust suboptimal mpc for autonomous racing with anytime feasible sqp,'' \emph{IEEE Robotics and Automation Letters}, 2024.

\end{thebibliography}

\end{document}